\documentclass{article}
\usepackage{microtype}
\usepackage{graphicx}
\usepackage{subcaption}
\usepackage{booktabs}
\usepackage{hyperref}

\usepackage[accepted]{icml2026}

\usepackage{amsmath}
\usepackage{amssymb}
\usepackage{mathtools}
\usepackage{amsthm}

\usepackage[capitalize,noabbrev]{cleveref}

\theoremstyle{plain}

\theoremstyle{definition}

\theoremstyle{remark}

\usepackage[textsize=tiny]{todonotes}

\icmltitlerunning{Targeted Parameter Decomposition}

\pdfoutput=1

\begin{document}

\twocolumn[
  \icmltitle{Targeted Recovery of Weight-Space Mechanisms From Neural Networks}

  \icmlsetsymbol{equal}{*}

  \begin{icmlauthorlist}
    \icmlauthor{Antoine Vigouroux}{av}
    \icmlauthor{Lee Sharkey}{ls}
  \end{icmlauthorlist}

  \icmlaffiliation{av}{MATS}
  \icmlaffiliation{ls}{Goodfire}

  \icmlcorrespondingauthor{Antoine Vigouroux}{antvig@pm.me}

  \icmlkeywords{Mechanistic Interpretability, Parameter Decomposition, Circuits, Language Models}

  \vskip 0.3in
]

\printAffiliationsAndNotice{}

\begin{abstract}
Parameter decomposition (PD) decomposes neural networks into interpretable computational components
that faithfully reflect the original network's operations.
However, scaling PD to large models requires vast compute,
making it a costly and risky endeavor. 
Here we propose targeted PD (tPD), 
which identifies only the components that process specific inputs of interest
-- from isolated prompts to large subtasks -- 
by introducing a high-rank catch-all component that handles all non-target data.
We validate tPD on toy models and on transformer language models trained on The Pile, 
where it recovers reproducible, 
mechanistically faithful circuits.
We extract a CSS-only submodel of a 4-block transformer using $\approx$7\% of the FLOPs of its published decomposition, 
and in a 12-block transformer we surgically ablate and rewire memorized sequences,
with negligible side effects on other inputs. 
\end{abstract}

\section{Introduction}

Parameter decomposition (PD) aims to reverse-engineer neural networks by recovering the algorithms they implement, rather than annotating their internal representations \citep{braun_interpretability_2025,bushnaq_stochastic_2025,bushnaq_interpreting_2026}.
Each weight matrix is decomposed into mechanisms made of rank-1 subcomponents $U_c V_c^\top$, each implementing a single functional role.
The decomposition must satisfy four requirements:
(i) the subcomponents sum to the original weights,
(ii) inactive subcomponents can be ablated in any combination without changing the model's outputs,
(iii) each subcomponent is as computationally simple as possible,
and (iv) each input activates a minimal number of components \citep{bushnaq_stochastic_2025,bushnaq_interpreting_2026}.
A jointly-trained auxiliary network learns which subcomponents are active (i.e., causally-important) on which inputs.
After initial work on toy models, this intricate optimization problem was recently scaled up to a 4-block transformer with 28M non-embedding parameters \citep{bushnaq_interpreting_2026}.
Reaching frontier LLMs would require several orders of magnitude more scale, with no ground truth to guide the engineering.

Here we show that the PD framework does not require full-model decomposition to be useful.
We introduce \textbf{Targeted Parameter Decomposition (tPD)},
which selectively identifies the mechanisms that process a chosen subset of data (the \textit{target data}),
substantially reducing the compute requirements.

\begin{figure*}
    \centering
    \includegraphics[width=\linewidth]{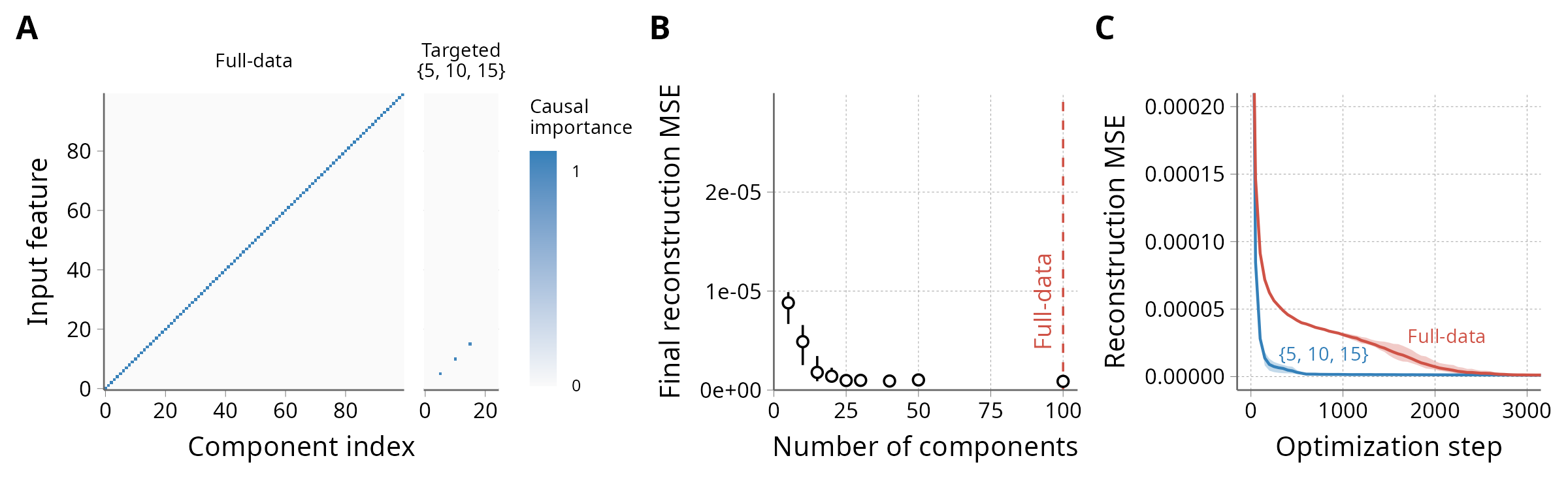}
    \includegraphics[width=\linewidth]{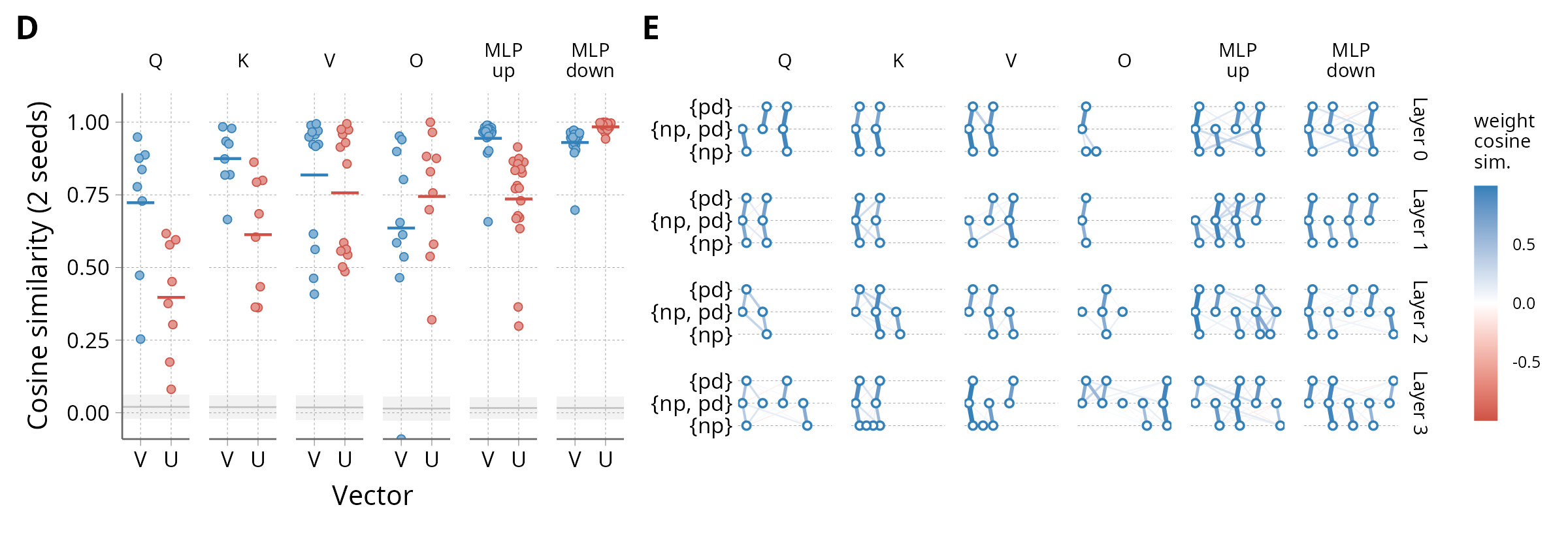}    
    \caption{
        \textbf{A:} Causal importance patterns for the TMCC toy model's first matrix: full-data and targeted on 3 features.
        \textbf{B:} Mean-square error (MSE) of the reconstruction, as a proxy for decomposition success, for targeted decomposition of the TMCC toy model when a variable number of subcomponent slots are provided.
        \textbf{C:} MSE over training for the full-data and targeted decompositions of the TMCC.
        \textbf{D:} Cosine similarities of the $U$ and $V$ vectors between matched components across two random seeds for the joint numpy/pandas task. 
            The horizontal bars represent the mean. 
            The gray line and shaded area represent the mean ± standard deviation of a null distribution obtained by comparing the first decomposition to a untrained, randomly-initialized version of the second one.
            For comparison, full data decompositions of the same model with different seeds \citep{bushnaq_interpreting_2026} yielded Mean Max Cosine Similarities of 0.48 and 0.51 for $U$ and $V$ vectors respectively.
        \textbf{E:} Pair-wise matches between the subcomponents obtained from decomposition on nested target sets: 
            \textit{import numpy as} (\{np\}), 
            \textit{import pandas as} (\{pd\}),
            or both at the same time (\{np, pd\}).
    }
    \label{fig:benchmarks}
\end{figure*}

\section{Method}
\label{sec:method}

A natural starting point would be to run PD using only the target data.
This recovers components that correctly reconstruct the model's outputs on the target \citep{christensen_decomposition_2025},
but leaves the solution underdetermined:
On a narrow target dataset,
activations span only a subspace of what the model sees on general data.
Components are constrained inside that subspace,
but are arbitrary in orthogonal directions.
Thus, their ablations would have unpredictable effects on non-target inputs --
hindering model editing and preventing us from inspecting the broader contexts in which they fire.

To obtain interpretable mechanisms,
tPD trains PD on two parallel data streams --
a smaller \textit{target} dataset of inputs we want to explain, and a \textit{non-target} stream sampled from the model's original training distribution.
The decomposition model is trained to reconstruct the model's outputs on both streams:
\begin{itemize}
    \item For target data, the reconstruction must be done using only a sparse set of rank-1 subcomponents, like in standard PD. However, these subcomponents are not required to sum to the original weights -- they only need to capture the mechanisms that fire on the target data.
    \item For non-target data, each weight matrix is allocated a full-rank, catch-all component $\Delta$, defined as the residual between the original weights and the sum of subcomponents. These $\Delta$ components capture all mechanisms that are never active on the target data, so they are not explicitly decomposed. The subcomponents that process target data can also be used for non-target reconstruction, since some mechanisms may be active on both target and non-target data.
\end{itemize}
To achieve this, the $\Delta$ components are adversarially ablated on target batches (each between 0 and 100\%) to maximize reconstruction loss,
forcing the target outputs to be reconstructed from the subcomponents alone.
On non-target batches, the $\Delta$ components are always fully enabled.

Importantly, both the target and non-target streams contribute to the importance-minimality loss.
This has two consequences:
(i) any mechanism that fires on non-target but not on target data is moved into the $\Delta$ components (which are not counted in the importance-minimality loss);
(ii) the subcomponents that \textit{do} process target data are shaped to interfere as little as possible with non-target data.

To ensure that the decomposed circuits are mechanistically faithful to the original computation,
both the inactive and $\Delta$ components are adversarially ablated to maximize reconstruction loss,
forcing the model's outputs to remain unchanged for any hybrid of original and ablated matrices.
See Appendix \ref{app:method_details} for more detail on the implementation, and Appendix \ref{app:tms} for a full worked example on the ``TMS-5-2-id" toy model \citep{bushnaq_stochastic_2025}, illustrating why the non-target batches and the $\Delta$ components are needed to capture the correct mechanisms.

\section{Results}

\subsection{tPD rapidly recovers ground-truth on toy models}

We first validate tPD on toy models.
To demonstrate the compute gains, we use the Toy Model of Compressed Computation (TMCC) from \citet{braun_interpretability_2025}.
In this model, parameter decomposition identifies 100 mechanisms for the MLP input matrix: One for each input feature (Fig \ref{fig:benchmarks}A, left).
We run the targeted version using only the inputs 5, 10 and 15 as the target data.
As expected, we recover exactly the three mechanisms that process these features (Fig \ref{fig:benchmarks}A, right).
Crucially, the CI network learns the correct activation patterns, so that the subcomponents do not spuriously activate on non-target inputs.
The computational cost is reduced for two reasons:
\begin{itemize}
    \item First, the decomposition model only needs to be large enough to accommodate the target mechanisms, for a much reduced memory footprint.
While the full TMCC requires at least 100 subcomponent slots per matrix, tPD on a subset of 3 features works with much smaller models (as few as 5 subcomponent slots, Fig. \ref{fig:benchmarks}B).
    \item Second, we find that tPD converges faster, presumably because it is a much simpler optimization problem.
All else equal, full-data PD on the 1-layer TMCC takes $\sim$2500 optimization steps to converge; tPD on a 3-input target converges in $\sim$500 (Fig. \ref{fig:benchmarks}C).
\end{itemize}

\subsection{Consistency checks on a transformer}

We next validate the approach on transformer language models, whose training data is far more complex than that of toy models.
Although there is no ground truth to compare against, we run two consistency checks:
First, if tPD discovers identifiable mechanisms, decompositions initialized from different random seeds should converge to similar circuits.
Second, decompositions of telescoping sets should be coherent: The circuits that process $A \cup B$ should be the union of the circuits that process $A$ and those that process $B$ (with possible exceptions due to higher-rank mechanisms, see Appendices \ref{app:tms}-\ref{app:different_components}).

We perform this validation on the largest model for which a full-data decomposition exists so far: A 4-block transformer trained on The Pile \citep{bushnaq_interpreting_2026}.
To validate tPD on the extreme opposite case, we decompose it against just two inputs: ``\texttt{import numpy as}" and ``\texttt{import pandas as}".
These sequences are omnipresent in Python code and the decomposed model correctly completes them as ``\texttt{ np}'' and ``\texttt{ pd}''.

Using these two prompts as target and the general Pile as non-target, we perform tPD twice, initializing from two different random seeds.
We then match subcomponents based on maximum cosine similarity between their weights and compute the cosine similarities between their respective $U$ and $V$ vectors (Fig. \ref{fig:benchmarks}D).
Overall, there is strong agreement.
The $V$ vectors (input directions) in particular are highly similar, suggesting that the two decompositions converged to essentially the same implementation.
The $U$ vectors in the Q and K matrices of the attention layers are slightly less consistent, indicating that the representations used to form the attention pattern are somewhat less constrained.
Note that there are multiple reasons why the subcomponents may or may not be the same across initializations.
See Appendix \ref{app:different_components} and Appendix B.3 from \citet{bushnaq_interpreting_2026} for further discussion.

To check for consistency across nested targets, we run targeted decompositions against each of the two prompts taken separately, and compare the subcomponents we obtain to those from the decomposition against both prompts at the same time.
Of the 77 subcomponents in the joint decomposition (\{np,pd\}), 69 have a strong match in either or both of the single-prompt decompositions: 18 are shared with the ``numpy'' decomposition (\{np\}), 19 with the ``pandas'' one (\{pd\}), and 32 are seemingly used by both prompts (Fig. \ref{fig:benchmarks}E).

\begin{figure}
    \centering
    \includegraphics[width=\linewidth]{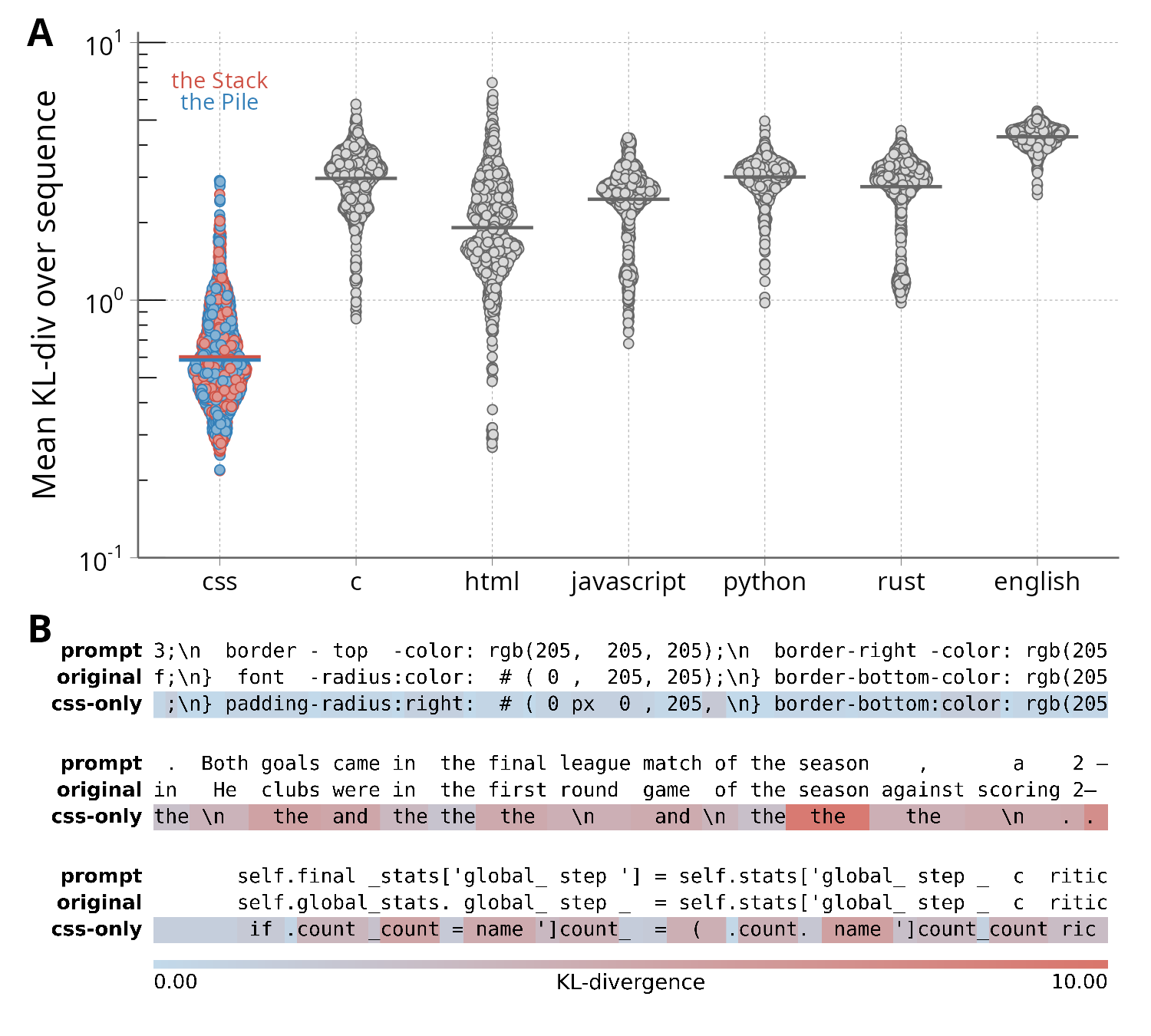}
    \caption{
        \textbf{A: } KL-divergence of a model made only of the subcomponents identified in the CSS-only decomposition, compared to the original model, 
            on various languages. Lower values mean the outputs are closer to the original model's.
        \textbf{B: } Comparison of the original sequence, next tokens predicted by the original model, 
            and next tokens predicted by the CSS-only model, on samples of text from CSS code, Python code, and English language. 
            The prompt line is shifted left by one position so the real next token is aligned to the predicted next token.
            Highlights are the per-token KL-divergence between the original and CSS-only models clamped to $[0,1]$.
    }
    \label{fig:css_only}
\end{figure}

\begin{figure*}
    \centering
    \includegraphics[width=\linewidth]{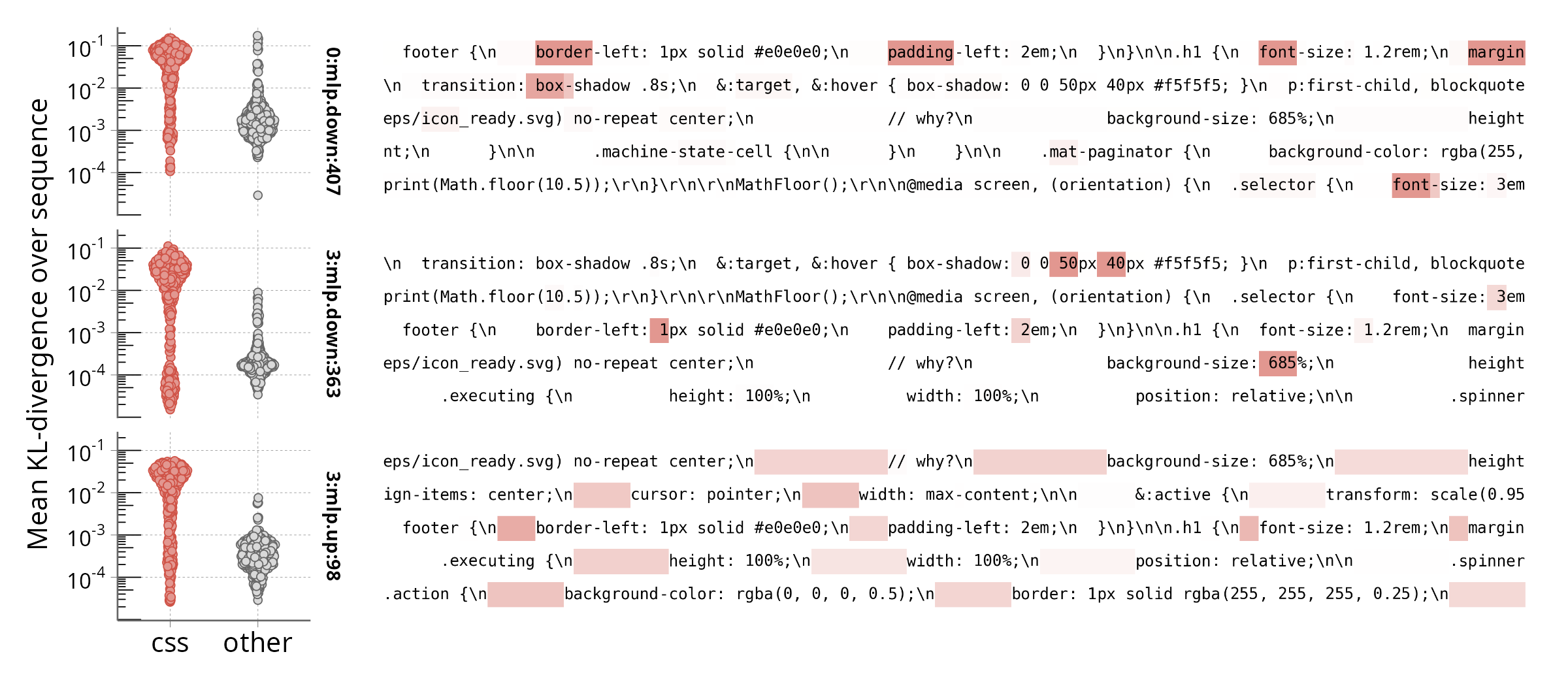}
    \caption{
        Three examples of CSS-specific subcomponents from the CSS-only decomposition. 
        \textbf{Left:} KL-divergence compared to the original model's output when the subcomponent is individually subtracted from the model, for CSS code versus other languages. 
        \textbf{Right:} Activation patterns of these components (displayed as the per-token KL-divergence when ablated, clamped to $[0;1]$).
    }
    \label{fig:shortlisted}
\end{figure*}

\subsection{Extracting a CSS-only submodel}

Having established that tPD produces consistent decompositions, we turn to a broader case: extracting every mechanism that processes CSS code.
Using the same 4-block transformer, we take as target a subset of the Pile restricted to CSS code, with comments stripped out to avoid picking up plain-language mechanisms.

Using $\approx$7\% of the FLOPs of the published full-data decomposition (Appendix \ref{app:flops}), we match and exceed its reconstruction accuracy (mean adversarial KL-divergence: 0.45, causal-importance L0-norm: 89.6).
tPD returns 1,638 subcomponents, against 9,972 for the full-data decomposition.
These 1,638 subcomponents presumably capture the algorithm the model uses to predict CSS code.
To verify this, we run a ``CSS-only'' model with only these components
on a panel of languages from The Stack \citep{kocetkov_stack_2022} and WikiText \citep{merity_pointer_2016}.
As seen in Fig. \ref{fig:css_only}A, the CSS-only model behaves similarly to the original on CSS data (KL-div.~$\approx$~0.6).
By contrast, behavior on every other language collapses (KL-div.~$\geq$~2),
and the model tends to fall back to a single dominant token (``\texttt{the}'' for English, ``\texttt{count}'' for Python).

A purported benefit of performing targeted PD with non-target batches and $\Delta$ components is that the identified mechanisms should have no side effects on examples outside the target data, unless they are actively used in processing them.
If this is true, we should be able to selectively destroy CSS-specific abilities from the model, while retaining the rest.
This is not as straightforward, because many subcomponents that are active on CSS are also active in non-CSS contexts, so simply ablating all 1,638 is not an option.
We start by shortlisting subcomponents that fire frequently on CSS data but rarely do so on general Pile data.
Manual inspection of this shortlist identifies subcomponents with specific syntactic roles, such as numbers preceding a unit, or whitespace used for indentation (Figs. \ref{fig:app:css_eoa_407}, \ref{fig:app:css_eoa_363}, \ref{fig:app:css_eoa_98}).
Strikingly, they are highly specific to the CSS context: For example, subcomponent 98 from block 3's MLP-up is active on indentation in CSS code, but not indentation in Python code (Fig. \ref{fig:app:css_eoa_98}).
We conjecture that these subcomponents, beyond recognizing surface patterns, also encode some form of CSS-context awareness, although we leave a more mechanistic understanding of how this works to future research.
Accordingly, ablating these subcomponents degrades the model's output on CSS data, but leaves other languages unchanged (Fig. \ref{fig:shortlisted}, left).

\begin{figure*}[t]
    \centering
    \includegraphics[width=\linewidth]{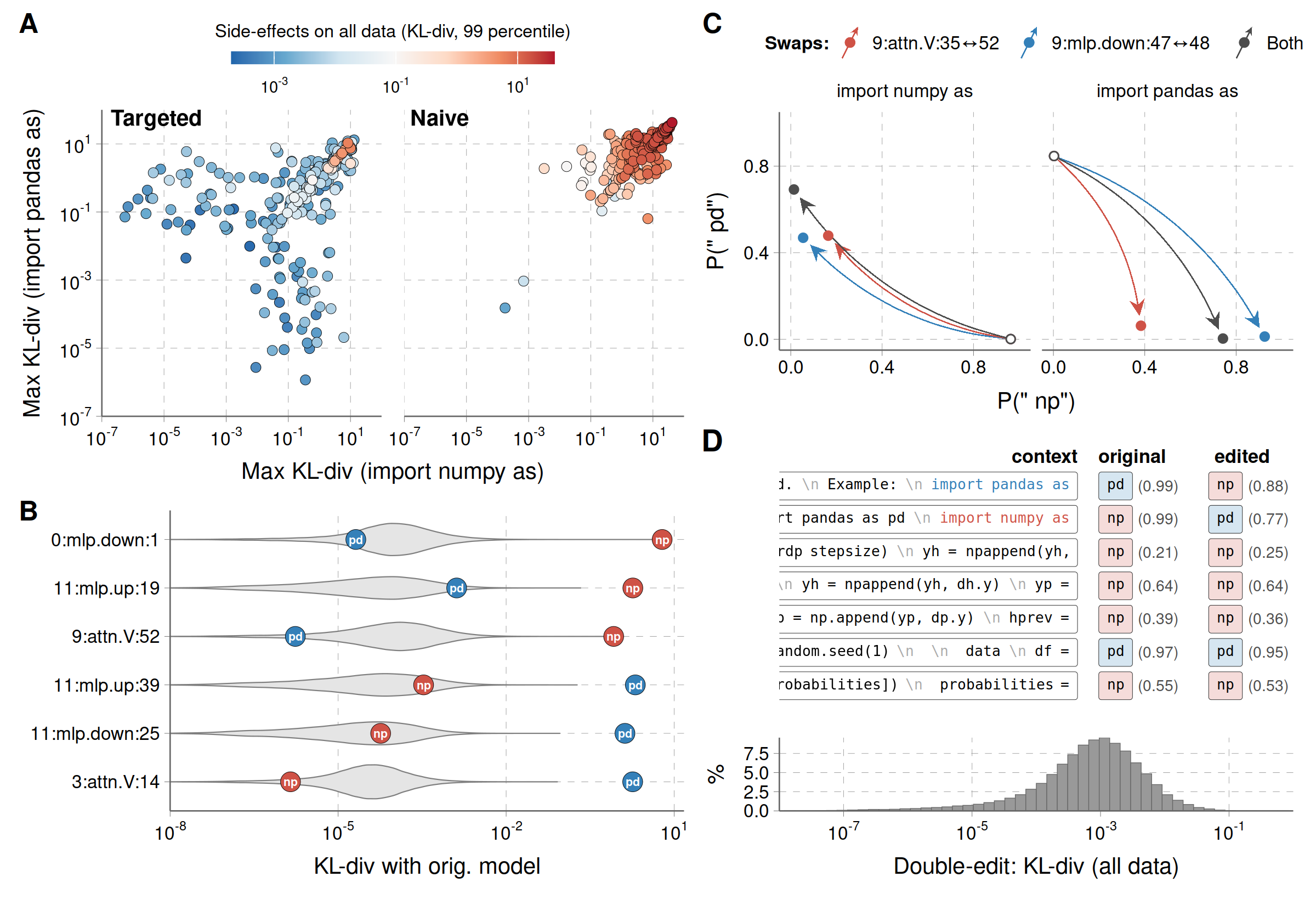}
    \caption{
        \textbf{A:} Effect of ablating individual subcomponents obtained either with the tPD method introduced here (``targeted"), or with standard PD on the target data without non-target batches (``naive").
        The x- and y-axes show the maximum $\text{KL}(\text{original}||\text{ablated})$ over the two target sequences. Point color represents the 99th-percentile KL-divergence on $>$100k tokens of general Pile data. Higher values mean the ablated model's output deviate farther from the original outputs.
        \textbf{B:} Erasing the np/pd completions from a 12-block model.
        The ``np" and ``pd" points show $\text{KL}(\text{original}||\text{ablated})$ on ``import numpy as" and ``import pandas as" respectively (higher values mean the ablation disrupts the model's behavior more on this input).
        Grey violins show the distribution of per-token divergences on general Pile data (lower values mean the ablation had less side effects). 
        \textbf{C:} Effect of swapping the output directions of components from the ``np" and ``pd" sequences on the predicted next token probabilities.
        \textbf{D:} Behavior of the swapped model on sample contexts, and distribution of divergences over general Pile data (lower means less side effects).
    }
    \label{fig:numpy}
\end{figure*}

\subsection{Editing knowledge in a 12-block transformer}
\label{sec:numpy}

A promising feature of tPD is that it lets us peek at the mechanisms inside larger models earlier,
before solving the engineering challenges required by full-data decomposition.
To demonstrate this, we come back to the ``numpy/pandas'' task above, and extend it to a 12-block Pile transformer (85M non-embedding parameters).
The decomposition finds 248 subcomponents (or about 5 per matrix).
Figure \ref{fig:numpy}A shows the effect of ablating each of them on the two target prompts.
When using our targeted decomposition scheme (left), the subcomponents fall into three broad categories: active only on the ``numpy'' prompt, active only on the ``pandas'' prompt, or active on both.
The remaining quadrant, active on neither ``np'' nor ``pd'', is empty:
In tPD, those are all absorbed into the $\Delta$ component.
We also quantify whether each subcomponent is used on non-target data
by calculating the ``worst-case'' (99th percentile) divergence on $>$100k tokens of generic Pile data,
excluding sequences that literally contain the target prompts.
Within the ``active on both'' category, some are also frequently active on non-target data (Fig. \ref{fig:numpy}A, red).
This includes, for example, subcomponents that activate at the beginning of every line.

The off-diagonal subcomponents are all highly specific to the numpy/np and pandas/pd completions.
Presumably, these subcomponents represent where these associations are stored in the model.
We can then ablate them to erase that knowledge from the model, completely suppressing the ability to complete either ``np'' or ``pd'', while retaining virtually identical behavior on other sequences (KL $<$ 10$^{-2}$, Fig. \ref{fig:numpy}B).
For comparison, we also run a naive decomposition (target-batch only, no $\Delta$ components) against these two prompts. 
In that case, the subcomponents all have heavy side effects on non-target data when ablated, despite correctly reconstructing the target data (Fig. \ref{fig:numpy}A, right).

Since each subcomponent is rank-1, it reads from a direction $V$ in activation space and writes to a different direction $U$.
We should therefore be able not only to ablate these connections, but to rewire them -- 
e.g., make the model import numpy as ``pd'' and pandas as ``np'' 
by swapping the $U$ directions of subcomponents specific to each prompt.
This is a stricter test of mechanistic faithfulness, because it is not something the decomposition explicitly optimizes for.

We select two pairs of highly prompt-specific subcomponents that seem like promising candidates for swapping \textit{a priori}\footnote{We only attempted the swap experiment on these two pairs.}:
\begin{itemize}
    \item Subcomponents 35 and 52 in the V projection of block 9's attention (which copy information from the second-to-last to the last token);
    \item Subcomponents 47 and 48 from block 9's MLP down-projection.
\end{itemize}
We then generate edited models where the output directions $U$ are swapped between the two subcomponents and scaled according to the ratio of their inner activations $V^\top x$,
while the input directions $V$ are left alone.
As predicted, the swapped models now complete \texttt{import numpy as} with \texttt{ pd} and \texttt{import pandas as} with \texttt{ np}, including when these sequences occur in natural code contexts (Fig. \ref{fig:numpy}C).
The change in behavior is remarkably specific: The completions are essentially unchanged (KL $<$ 10$^{-1}$) on most tokens from the Pile data, including on positions where the model initially predicts `` np'' or `` pd'' in unrelated contexts (Fig. \ref{fig:numpy}D).

\section{Discussion}

tPD is a targeted variant of PD that decomposes only the mechanisms
that are causally important
on a chosen set of inputs.
Our experiments suggest that tPD recovers identifiable, mechanistically faithful decompositions that support targeted interventions -- component ablation and rewiring of individual rank-1 connections -- at a small fraction of the cost of full-data PD.
This makes it a practical tool for narrow-scope interpretability \citep{sharkey_open_2025}: Circuit localization, unlearning, and editing at scales where full-data PD might be challenging.
It should also help scale full-data PD itself, by speeding up experimental iteration and surfacing facts about the target models that can serve as anchors for tuning the full decomposition.

However, several caveats apply.
First, the subcomponents from tPD only process the slice of activation space that is spanned by target data -- therefore, they are not directly identical to the full-data subcomponents (Appendix \ref{app:different_components}).
Depending on the application, it will be important to choose the target data such that it fully captures the behavior of interest.
Second, our editing case study (numpy/pandas, section \ref{sec:numpy}) is a relatively easy target for unlearning (short, token-specific memorized sequences).
Editing world knowledge out of LLMs will likely require more involved engineering,
both in choosing the target data and in designing the edits.
Third, neural networks sometimes implement one task with several redundant mechanisms \citep{mcgrath_hydra_2023,wang_interpretability_2022,bushnaq_interpreting_2026}.
This could cause some mechanisms to be lost in the decomposition, as the original model's behavior could be recreated using only a fraction of the relevant mechanisms.
This concern has not yet been addressed for full-data PD either, but it may be sharper in tPD
where inactive mechanisms are absorbed by the $\Delta$ components.
We provide a preliminary exploration of this in Appendix \ref{app:completeness}, but the question is far from settled.
Finally, even the largest model used here (similar in size to GPT-2-small) is still tiny by frontier-model standards, and scaling tPD further may surface unforeseen challenges.

\section{Related work}
\label{related_work}

\paragraph{Parameter decomposition.}
Parameter decomposition was introduced by APD \citep{braun_interpretability_2025}, which framed mechanistic interpretation as recovering an additive set of full-rank weight subcomponents, each implementing a single functional role.
SPD \citep{bushnaq_stochastic_2025} reformulated the optimization with rank-1 subcomponents and stochastic ablations, making the framework tractable on harder toy models;
VPD \citep{bushnaq_interpreting_2026} added adversarial ablations and a simplicity criterion, and was the first variant to scale to full decomposition of a transformer language model.
All three are \textit{full-data} methods: The recovered subcomponents are required to sum to the original weights, and the decomposition is trained on a dataset meant to be representative of the model's full input distribution.

The closest neighbor to tPD is \citet{christensen_decomposition_2025}, who run unmodified PD on isolated prompts.
The decomposition then reconstructs outputs on the target, but, as discussed in Section~\ref{sec:method}, the resulting decomposition is underdetermined off-target.

\paragraph{Targeted interpretability on data subsets.}
Most unsupervised mechanistic-interpretability methods such as sparse autoencoders (SAEs, \citet{cunningham_sparse_2023}), transcoders \citep{dunefsky_transcoders_2024} and Lorsa \citep{he_towards_2025} are run on the model's full input distribution, with the goal of producing a complete explanation of the model.
A handful of recent works trained SAEs on a restricted domain rather than on broad data:
\citet{muhamed_decoding_2024} introduce {specialized} SAEs trained on retrieval-selected subdomains, to surface rare concepts that broad-distribution SAEs miss.
\citet{oneill_resurrecting_2025} confine SAE training to a single domain (clinical text), reallocating dictionary capacity to domain-specific features.
In both cases, the aim is to improve SAE performance on the target data, rather than reduce the compute requirements.
They inherit the concern that activation-space methods may not capture all of the causally relevant structure carried by the weights \citep{bilalchughtai_activation_2025}.

\paragraph{Circuit discovery.}
A second line of work explains how a model implements a pre-defined task in a \emph{supervised} manner.
For example, activation patching \citep{heimersheim_how_2024} localizes the components responsible for the difference between a clean and a corrupted prompt;
ACDC \citep{conmy_towards_2023} automates this into a search over the activation-space computational graph.

In contrast, tPD is fundamentally \textit{unsupervised}: the user supplies a slice of inputs but no behavioral metric and no assumption about the nature of the task.
The decomposition surfaces the mechanisms that are causally important on those inputs, and it is then up to the researcher to ``read the task" off the recovered components rather than specify it in advance.

\paragraph{Knowledge editing and unlearning.}

Diverse methods have been developed to locate and edit factual knowledge in language models.
For example, a major class of methods edits the matrices of MLP modules, treating them as look-up tables for factual associations \citep{meng_locating_2023,meng_mass-editing_2023}.
Further developments improved on the locality of edits (especially under sequential editing), by projecting the update on the nullspace of preserved knowledge \citep{fang_alphaedit_2025,lyu_evoedit_2026} or by localizing edits to individual MLP neurons \citep{pan_precise_2025}.
Another approach is to learn a hypernetwork that generates weight updates encouraging desired input/output pairs \citep{mitchell_fast_2022,li_reinforced_2025}.
Other methods such as REMEDI \citep{hernandez_inspecting_2024} and SAKE \citep{scialanga_sake_2025} edit the network’s representations rather than the weights.

These methods nonetheless start from a researcher-curated definition of the fact to be edited — a target input/output pair \citep{mitchell_fast_2022}, a subject-relation-object triple \citep{meng_locating_2023,meng_mass-editing_2023}, a counterfactual corpus \citep{eldan_whos_2023} or a distribution of paraphrases and implications \citep{scialanga_sake_2025}.
Moreover, several of them presuppose that the relevant knowledge is stored in a particular form, typically as key-value associations in specific MLP layers \citep{geva_transformer_2021}.
Parameter decomposition makes neither assumption: It surfaces the mechanisms causally responsible for a chosen set of inputs directly from the weights, without assuming what is stored or how.
While we demonstrate only edits on simple memorized sequences, we see this as a first step toward mechanistically-informed editing, in which edits are designed around a reverse-engineered mechanism rather than imposed on the output.


\section*{Acknowledgments}
Thanks to Cory Kendrick and Tasos Spiliotopoulos for support and feedback throughout the project. Thanks to Dan Braun for feedback on this manuscript and for training the 12-block transformer model used here. This work was supported by MATS.

\section*{Code availability}
Source code for this project can be found at:\\
\url{https://github.com/Antovigo/targeted-parameter-decomposition}.

\bibliography{references}

\begin{thebibliography}{28}
\providecommand{\natexlab}[1]{#1}
\providecommand{\url}[1]{\texttt{#1}}
\expandafter\ifx\csname urlstyle\endcsname\relax
  \providecommand{\doi}[1]{doi: #1}\else
  \providecommand{\doi}{doi: \begingroup \urlstyle{rm}\Url}\fi

\bibitem[bilalchughtai \& Bushnaq(2025)bilalchughtai and Bushnaq]{bilalchughtai_activation_2025}
bilalchughtai and Bushnaq, L.
\newblock Activation space interpretability may be doomed — {LessWrong}.
\newblock January 2025.

\bibitem[Braun et~al.(2025)Braun, Bushnaq, Heimersheim, Mendel, and Sharkey]{braun_interpretability_2025}
Braun, D., Bushnaq, L., Heimersheim, S., Mendel, J., and Sharkey, L.
\newblock Interpretability in {Parameter} {Space}: {Minimizing} {Mechanistic} {Description} {Length} with {Attribution}-based {Parameter} {Decomposition}, February 2025.
\newblock URL \url{http://arxiv.org/abs/2501.14926}.
\newblock arXiv:2501.14926 [cs].

\bibitem[Bushnaq et~al.(2026)Bushnaq, Braun, {*}, Clive-Griffin, {*}, and Sharkey]{bushnaq_interpreting_2026}
Bushnaq, L., Braun, D., {*}, Clive-Griffin, O., {*}, and Sharkey, L.
\newblock Interpreting {Language} {Model} {Parameters}, 2026.
\newblock URL \url{https://www.goodfire.ai/research/interpreting-lm-parameters}.

\bibitem[Bushnaq et~al.(2025)Bushnaq, Braun, and Sharkey]{bushnaq_stochastic_2025}
Bushnaq, L., Braun, D., and Sharkey, L.
\newblock Stochastic {Parameter} {Decomposition}, September 2025.
\newblock URL \url{http://arxiv.org/abs/2506.20790}.
\newblock arXiv:2506.20790 [cs].

\bibitem[Christensen \& Riggs(2025)Christensen and Riggs]{christensen_decomposition_2025}
Christensen, C.~L. and Riggs, L.
\newblock Decomposition of {Small} {Transformer} {Models}, December 2025.
\newblock URL \url{http://arxiv.org/abs/2511.08854}.
\newblock arXiv:2511.08854 [cs].

\bibitem[Conmy et~al.(2023)Conmy, Mavor-Parker, Lynch, Heimersheim, and Garriga-Alonso]{conmy_towards_2023}
Conmy, A., Mavor-Parker, A.~N., Lynch, A., Heimersheim, S., and Garriga-Alonso, A.
\newblock Towards {Automated} {Circuit} {Discovery} for {Mechanistic} {Interpretability}, October 2023.
\newblock URL \url{http://arxiv.org/abs/2304.14997}.
\newblock arXiv:2304.14997 [cs].

\bibitem[Cunningham et~al.(2023)Cunningham, Ewart, Riggs, Huben, and Sharkey]{cunningham_sparse_2023}
Cunningham, H., Ewart, A., Riggs, L., Huben, R., and Sharkey, L.
\newblock Sparse {Autoencoders} {Find} {Highly} {Interpretable} {Features} in {Language} {Models}, October 2023.
\newblock URL \url{http://arxiv.org/abs/2309.08600}.
\newblock arXiv:2309.08600 [cs].

\bibitem[Dunefsky et~al.(2024)Dunefsky, Chlenski, and Nanda]{dunefsky_transcoders_2024}
Dunefsky, J., Chlenski, P., and Nanda, N.
\newblock Transcoders {Find} {Interpretable} {LLM} {Feature} {Circuits}, November 2024.
\newblock URL \url{http://arxiv.org/abs/2406.11944}.
\newblock arXiv:2406.11944 [cs].

\bibitem[Eldan \& Russinovich(2023)Eldan and Russinovich]{eldan_whos_2023}
Eldan, R. and Russinovich, M.
\newblock Who's {Harry} {Potter}? {Approximate} {Unlearning} in {LLMs}, October 2023.
\newblock URL \url{http://arxiv.org/abs/2310.02238}.
\newblock arXiv:2310.02238 [cs].

\bibitem[Fang et~al.(2025)Fang, Jiang, Wang, Ma, Jie, Wang, He, and Chua]{fang_alphaedit_2025}
Fang, J., Jiang, H., Wang, K., Ma, Y., Jie, S., Wang, X., He, X., and Chua, T.-s.
\newblock {AlphaEdit}: {Null}-{Space} {Constrained} {Knowledge} {Editing} for {Language} {Models}, April 2025.
\newblock URL \url{http://arxiv.org/abs/2410.02355}.
\newblock arXiv:2410.02355 [cs.CL].

\bibitem[Geva et~al.(2021)Geva, Schuster, Berant, and Levy]{geva_transformer_2021}
Geva, M., Schuster, R., Berant, J., and Levy, O.
\newblock Transformer {Feed}-{Forward} {Layers} {Are} {Key}-{Value} {Memories}, September 2021.
\newblock URL \url{http://arxiv.org/abs/2012.14913}.
\newblock arXiv:2012.14913 [cs.CL].

\bibitem[He et~al.(2025)He, Wang, Lin, Ge, Shu, Tang, Zhang, and Qiu]{he_towards_2025}
He, Z., Wang, J., Lin, R., Ge, X., Shu, W., Tang, Q., Zhang, J., and Qiu, X.
\newblock Towards {Understanding} the {Nature} of {Attention} with {Low}-{Rank} {Sparse} {Decomposition}, April 2025.
\newblock URL \url{http://arxiv.org/abs/2504.20938}.
\newblock arXiv:2504.20938 [cs].

\bibitem[Heimersheim \& Nanda(2024)Heimersheim and Nanda]{heimersheim_how_2024}
Heimersheim, S. and Nanda, N.
\newblock How to use and interpret activation patching, April 2024.
\newblock URL \url{http://arxiv.org/abs/2404.15255}.
\newblock arXiv:2404.15255 [cs].

\bibitem[Hernandez et~al.(2024)Hernandez, Li, and Andreas]{hernandez_inspecting_2024}
Hernandez, E., Li, B.~Z., and Andreas, J.
\newblock Inspecting and {Editing} {Knowledge} {Representations} in {Language} {Models}, August 2024.
\newblock URL \url{http://arxiv.org/abs/2304.00740}.
\newblock arXiv:2304.00740 [cs.CL].

\bibitem[Kocetkov et~al.(2022)Kocetkov, Li, Allal, Li, Mou, Ferrandis, Jernite, Mitchell, Hughes, Wolf, Bahdanau, Werra, and Vries]{kocetkov_stack_2022}
Kocetkov, D., Li, R., Allal, L.~B., Li, J., Mou, C., Ferrandis, C.~M., Jernite, Y., Mitchell, M., Hughes, S., Wolf, T., Bahdanau, D., Werra, L.~v., and Vries, H.~d.
\newblock The {Stack}: 3 {TB} of permissively licensed source code, November 2022.
\newblock URL \url{http://arxiv.org/abs/2211.15533}.
\newblock arXiv:2211.15533 [cs].

\bibitem[Li et~al.(2025)Li, Jiang, Chen, Bi, Zhou, Sun, Fang, and Wang]{li_reinforced_2025}
Li, Z., Jiang, H., Chen, H., Bi, B., Zhou, Z., Sun, F., Fang, J., and Wang, X.
\newblock Reinforced {Lifelong} {Editing} for {Language} {Models}, September 2025.
\newblock URL \url{http://arxiv.org/abs/2502.05759}.
\newblock arXiv:2502.05759 [cs.CL].

\bibitem[Lyu et~al.(2026)Lyu, Gu, Wang, Huang, Luan, Cui, Chang, and Lu]{lyu_evoedit_2026}
Lyu, S., Gu, Y., Wang, X., Huang, J., Luan, S., Cui, Y., Chang, X.-W., and Lu, P.
\newblock {EvoEdit}: {Evolving} {Null}-space {Alignment} for {Robust} and {Efficient} {Knowledge} {Editing}, April 2026.
\newblock URL \url{http://arxiv.org/abs/2510.13851}.
\newblock arXiv:2510.13851 [cs.CL].

\bibitem[McGrath et~al.(2023)McGrath, Rahtz, Kramar, Mikulik, and Legg]{mcgrath_hydra_2023}
McGrath, T., Rahtz, M., Kramar, J., Mikulik, V., and Legg, S.
\newblock The {Hydra} {Effect}: {Emergent} {Self}-repair in {Language} {Model} {Computations}, July 2023.
\newblock URL \url{http://arxiv.org/abs/2307.15771}.
\newblock arXiv:2307.15771 [cs].

\bibitem[Meng et~al.(2023{\natexlab{a}})Meng, Bau, Andonian, and Belinkov]{meng_locating_2023}
Meng, K., Bau, D., Andonian, A., and Belinkov, Y.
\newblock Locating and {Editing} {Factual} {Associations} in {GPT}, January 2023{\natexlab{a}}.
\newblock URL \url{http://arxiv.org/abs/2202.05262}.
\newblock arXiv:2202.05262 [cs].

\bibitem[Meng et~al.(2023{\natexlab{b}})Meng, Sharma, Andonian, Belinkov, and Bau]{meng_mass-editing_2023}
Meng, K., Sharma, A.~S., Andonian, A., Belinkov, Y., and Bau, D.
\newblock Mass-{Editing} {Memory} in a {Transformer}, August 2023{\natexlab{b}}.
\newblock URL \url{http://arxiv.org/abs/2210.07229}.
\newblock arXiv:2210.07229 [cs].

\bibitem[Merity et~al.(2016)Merity, Xiong, Bradbury, and Socher]{merity_pointer_2016}
Merity, S., Xiong, C., Bradbury, J., and Socher, R.
\newblock Pointer {Sentinel} {Mixture} {Models}, September 2016.
\newblock URL \url{http://arxiv.org/abs/1609.07843}.
\newblock arXiv:1609.07843 [cs].

\bibitem[Mitchell et~al.(2022)Mitchell, Lin, Bosselut, Finn, and Manning]{mitchell_fast_2022}
Mitchell, E., Lin, C., Bosselut, A., Finn, C., and Manning, C.~D.
\newblock Fast {Model} {Editing} at {Scale}, June 2022.
\newblock URL \url{http://arxiv.org/abs/2110.11309}.
\newblock arXiv:2110.11309 [cs].

\bibitem[Muhamed et~al.(2024)Muhamed, Diab, and Smith]{muhamed_decoding_2024}
Muhamed, A., Diab, M., and Smith, V.
\newblock Decoding {Dark} {Matter}: {Specialized} {Sparse} {Autoencoders} for {Interpreting} {Rare} {Concepts} in {Foundation} {Models}, November 2024.
\newblock URL \url{http://arxiv.org/abs/2411.00743}.
\newblock arXiv:2411.00743 [cs] version: 1.

\bibitem[O'Neill et~al.(2025)O'Neill, Jayasekara, and Kirkby]{oneill_resurrecting_2025}
O'Neill, C., Jayasekara, M., and Kirkby, M.
\newblock Resurrecting the {Salmon}: {Rethinking} {Mechanistic} {Interpretability} with {Domain}-{Specific} {Sparse} {Autoencoders}, August 2025.
\newblock URL \url{http://arxiv.org/abs/2508.09363}.
\newblock arXiv:2508.09363 [cs] version: 1.

\bibitem[Pan et~al.(2025)Pan, Wang, Cao, Shi, Yang, Li, and Wang]{pan_precise_2025}
Pan, H., Wang, X., Cao, Y., Shi, Z., Yang, X., Li, J., and Wang, M.
\newblock Precise {Localization} of {Memories}: {A} {Fine}-grained {Neuron}-level {Knowledge} {Editing} {Technique} for {LLMs}, March 2025.
\newblock URL \url{http://arxiv.org/abs/2503.01090}.
\newblock arXiv:2503.01090 [cs.CL] version: 2.

\bibitem[Scialanga et~al.(2025)Scialanga, Laugel, Grari, and Detyniecki]{scialanga_sake_2025}
Scialanga, M., Laugel, T., Grari, V., and Detyniecki, M.
\newblock {SAKE}: {Steering} {Activations} for {Knowledge} {Editing}, July 2025.
\newblock URL \url{http://arxiv.org/abs/2503.01751}.
\newblock arXiv:2503.01751 [cs.AI].

\bibitem[Sharkey et~al.(2025)Sharkey, Chughtai, Batson, Lindsey, Wu, Bushnaq, Goldowsky-Dill, Heimersheim, Ortega, Bloom, Biderman, Garriga-Alonso, Conmy, Nanda, Rumbelow, Wattenberg, Schoots, Miller, Michaud, Casper, Tegmark, Saunders, Bau, Todd, Geiger, Geva, Hoogland, Murfet, and McGrath]{sharkey_open_2025}
Sharkey, L., Chughtai, B., Batson, J., Lindsey, J., Wu, J., Bushnaq, L., Goldowsky-Dill, N., Heimersheim, S., Ortega, A., Bloom, J., Biderman, S., Garriga-Alonso, A., Conmy, A., Nanda, N., Rumbelow, J., Wattenberg, M., Schoots, N., Miller, J., Michaud, E.~J., Casper, S., Tegmark, M., Saunders, W., Bau, D., Todd, E., Geiger, A., Geva, M., Hoogland, J., Murfet, D., and McGrath, T.
\newblock Open {Problems} in {Mechanistic} {Interpretability}, January 2025.
\newblock URL \url{http://arxiv.org/abs/2501.16496}.
\newblock arXiv:2501.16496 [cs].

\bibitem[Wang et~al.(2022)Wang, Variengien, Conmy, Shlegeris, and Steinhardt]{wang_interpretability_2022}
Wang, K., Variengien, A., Conmy, A., Shlegeris, B., and Steinhardt, J.
\newblock Interpretability in the {Wild}: a {Circuit} for {Indirect} {Object} {Identification} in {GPT}-2 small, November 2022.
\newblock URL \url{http://arxiv.org/abs/2211.00593}.
\newblock arXiv:2211.00593 [cs].

\end{thebibliography}
\bibliographystyle{icml2026}

\newpage
\appendix
\onecolumn


\setcounter{figure}{0}
\renewcommand{\thefigure}{A\arabic{figure}}

\section{Method details}
\label{app:method_details}

In tPD, the decomposition model for each matrix can be written as:

\begin{equation}
W \approx W' = \sum_{i=1}^{C} m_i \, U_i V_i^\top + m_{\Delta} (W - \sum_{i=1}^{C} U_i V_i^\top),
\end{equation}

where $W$ is the original weight matrix, $V_i$ and $U_i$ are the input and output vectors of the rank-1 subcomponent $i$, and $C$ is the number of allocated subcomponent slots.
The subcomponent masks $m_i \in [\mu_i, 1]$ are bounded below by the causal importance $\mu_i$, itself determined by the auxiliary causal-importance network. 

\textbf{On target data}, the masks are chosen adversarially to maximize reconstruction loss:
\begin{equation}
m_1, \dots, m_C,m_\Delta = \text{argmax}_{m_i \in [\mu_i, 1]} \mathcal{L}_{\text{recon}}(W'),
\end{equation}
where $\mathcal{L}_{\text{recon}}(W')$ is the Kullback-Leibler divergence (KL) between the output probability distributions of the original model ($W$) and of the decomposition model ($W'$).
The adversarial ablation levels are set using Projected Gradient Descent with persistent, shared-across-batch adversarial sources \citep{bushnaq_interpreting_2026}.
In addition, to help with decomposition in the early stages of training, we add a stochastic reconstruction loss where the masks are sampled uniformly between $\mu_i$ and 1, as in \citet{bushnaq_stochastic_2025}.

\textbf{On non-target data}, $m_\Delta$ is always set to 1, and the subcomponent masks are sampled uniformly between $m_i$ and $\mu_i$.
For the CSS-only decomposition, we found it useful to add an \textit{unmasked} reconstruction loss, with all $m_i$ set to 1 and $m_\Delta$ set to 0, to prevent the dead components that never activate from interfering with the reconstruction. 
We also use a weight decay of 0.1 on subcomponent vectors, where the weight decay is scaled by $1 - \max_\text{batch}(\text{CI})$, such that the weights of dead components (that never get assigned a high causal importance CI over the batch) get dragged to zero.

In practice, we sample one batch of target data and one batch of non-target data, and run them sequentially. 
The gradients are accumulated over both batches before stepping the optimizer.
For all the experiments presented here, we use equal sizes for the target and non-target batches, though this could be optimized depending on the application.
To account for the fact that fewer subcomponents are active on the non-target batches, the importance-minimality coefficient is doubled for non-target batches.

\begin{figure}
    \centering
    \includegraphics[width=0.8\linewidth]{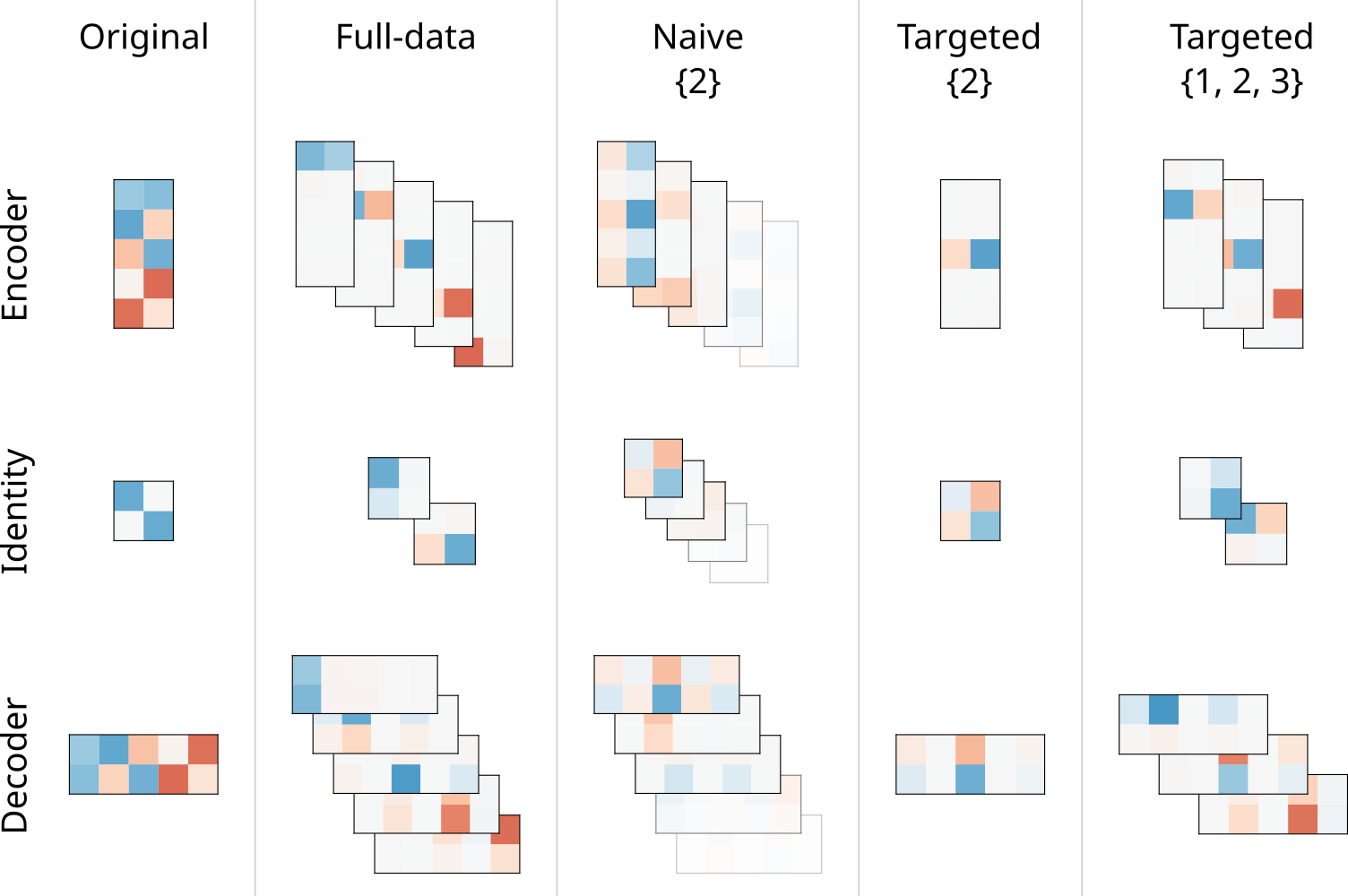}
    \caption{Weight matrices for the TMS 5-2-id toy model \citep{bushnaq_stochastic_2025}, and the subcomponents obtained after decomposition with full input data (5 features), with the original SPD algorithm but using \{2\} as the only input, with tPD using \{2\} as the only input, and with tPD using \{1, 2, 3\} as inputs. Red indicates positive values and blue negative values. The input dimension is shown vertically, and the output dimension horizontally.}
    \label{fig:app:tms}
\end{figure}

\section{Illustrative example: the TMS 5-2-id}
\label{app:tms}

Fig. \ref{fig:app:tms} illustrates the differences between naive PD and tPD using the TMS 5-2-id toy model from \citet{bushnaq_stochastic_2025}.
In this toy model, five sparsely-activating input dimensions are compressed into a 2D space. 
The 2D representation goes through an identity matrix (inserted in the toy model as a stand-in for a rank-2 mechanism), and decoded back to the original 5 dimensions. 
For the encoder and decoder, full-data PD recovers a single component for each input dimension.
The encoder components map each input feature to its hidden representation, and the decoder's components map them back to the original inputs, plus some superposition interference.
Meanwhile, the identity matrix is captured as two permanently-active rank-1 subcomponents.

Suppose we aim to recover the mechanisms that process input dimension 2. If we naively run PD using this dimension as the only input, 
the components we find are only accurate along the direction of the activations that occur in the target data. 
For other directions, they are completely unconstrained.
For example, in the ``Naive \{2\}'' decomposition (using input feature 2 as the only active input dimension), row \#2 of the encoder captures the network's operation, while the other rows are free to take arbitrary values.
Since these rows never receive any non-zero activations, their values have no effect.
Furthermore, the naive decomposition generates many dead components with non-zero weights in all rows except row \#2.

With the targeted decomposition scheme, we recover only one subcomponent for each matrix.
For the encoder and decoder, that subcomponent is similar to the one from the full-data decomposition.
For the identity matrix, the recovered subcomponent maps the hidden representation of feature \#2 to itself.
This is because, for higher-rank mechanisms, tPD only finds the slice that is actually spanned by the target data.

However, when two or more features are included in the target data (for example, when TMS-5-2-id is decomposed using \{1, 2, 3\} as the target), we recover the complete rank-2 identity matrix.

\section{When do full-data PD and targeted PD find different subcomponents?}
\label{app:different_components}

\begin{figure}
    \centering
    \includegraphics[width=.65\linewidth]{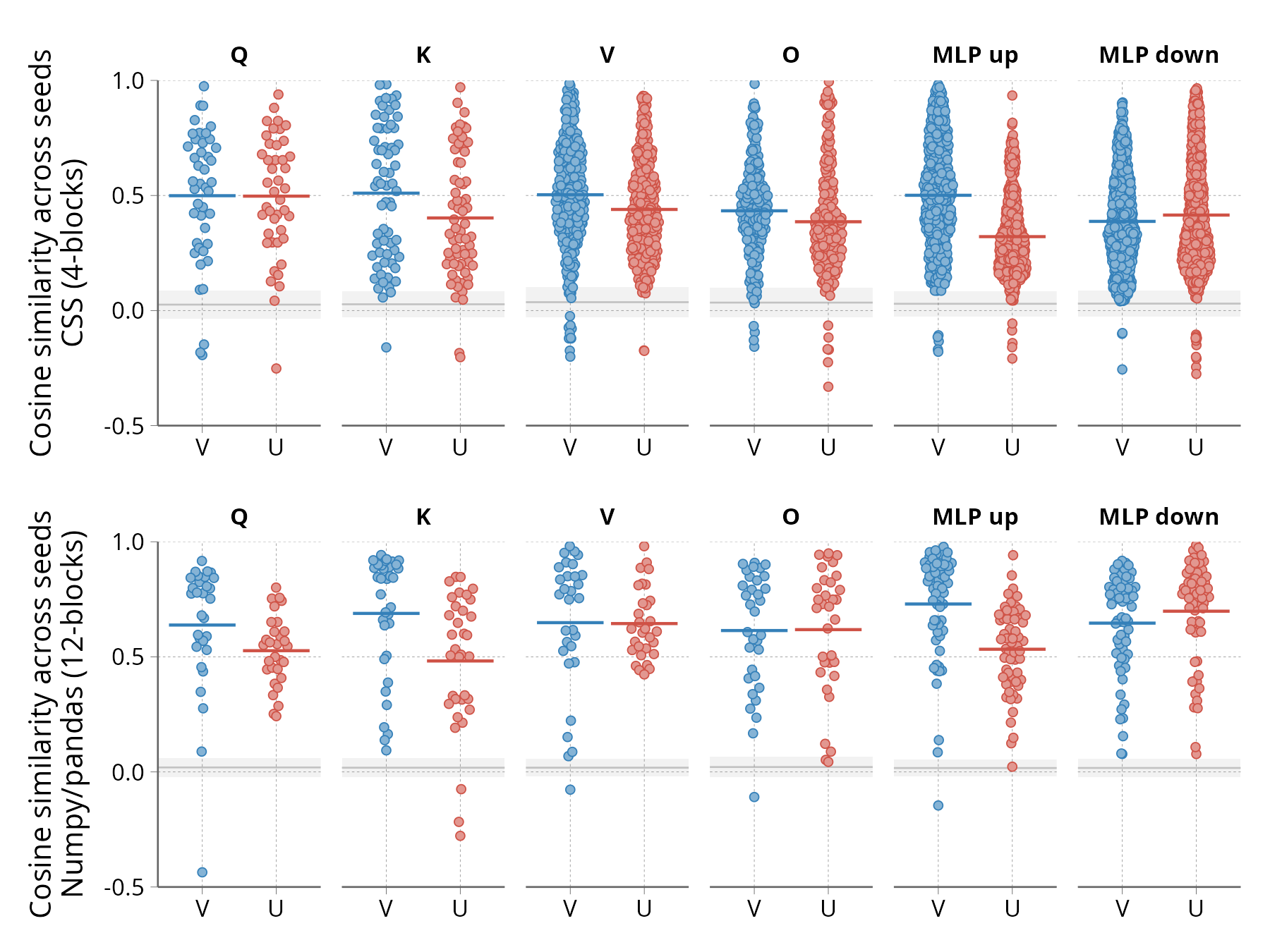}
    \caption{
        Cosine similarities between the $U$ and $V$ vector of matched subcomponents from two decompositions initialized with different random seeds.
        For each alive subcomponent in decomposition $a$, we find the closest alive subcomponent in decomposition $b$
        by computing the pairwise cosine similarities between the flattened matrices $UV^\top$ and taking the best match.\\
        Horizontal lines represent the mean.
        As a null distribution, the shaded gray areas represent the mean ± standard deviation of the cosine similarities obtained when decomposition $b$ is replaced by untrained, randomly-initialized values.
    }
    \label{fig:app:sim_across_seeds}
\end{figure}

When running targeted PD on a narrow dataset, the resulting subcomponents are not the same as the ones that would be discovered from a full-data decomposition.
In fact, even running the same decomposition twice with different random initializations is not expected to yield the exact same subcomponents every time (Fig. \ref{fig:app:sim_across_seeds}).
This can happen for multiple reasons, only some of which are undesirable:

\begin{enumerate}
    \item \textit{Unfaithful mechanisms:} the mechanisms discovered by PD are unrelated to the original model's computation -- instead, they are alternate implementations that simply output the same results. In PD and tPD, this is prevented by adversarial ablation of the inactive components.
    \item \textit{System drift:} the discovered mechanisms may be topologically correct,
 but deviate slightly from the original model in the exact feature vectors they use.
For example, the features written by the K and Q matrices of an attention block may both rotate compared to the ones used by the original, 
yet still ``recognize'' each other and leave behavior unchanged.
Adversarial ablations are meant to prevent this 
(the ablated K must still communicate with the original Q, and vice-versa), 
but we consistently observe that the U vectors in the Q/K matrices are less robust than the rest, 
indicating residual drift. 
Similar drift likely affects the U vectors of MLP up-projections and the V vectors of MLP down-projections.

    \item \textit{High-rank components:} some mechanisms are intrinsically higher-rank transformations. 
PD still captures them as multiple rank-1 subcomponents that simply sum to the weights of the complete mechanism \citep{bushnaq_stochastic_2025}.
The values of the individual subcomponents are then meaningless, 
as they only need to sum to the weights of the full mechanism, 
and there is no reason to expect them to match across initializations (see Fig. \ref{fig:app:tms} for an example).
    
    \item \textit{Subcomponent merging on narrow target data:} Suppose that two subcomponents from two distinct mechanisms reside in the same matrix, responding to different subspaces of activation-space.
In some cases, targeted decomposition may sum them as a single rank-1 subcomponent. 
For example, this occurs when two subcomponents always co-activate within the target data, with constant inner activations ($V^\top x$).
This is visible in our ``numpy/pandas" case study on the 4-block transformer, where we find only 77 alive subcomponents, 
while the same inputs activate no less than 1,708 subcomponents in the full-data decomposition of the same model.
        
    \item \textit{Decomposition error:} finally, some variation between subcomponents across runs may simply reflect each decomposition not fully converging within the allocated training steps.

\end{enumerate}


Overall, our results suggest that tPD produces reasonably faithful descriptions of the original model's computation. System drift and decomposition error likely occur, but not more than in full-data PD \citep{bushnaq_interpreting_2026}.
Component merging is a crucial aspect of targeted decomposition: in practice, the target data must be chosen so that it splits the components of interest at the desired granularity, while activating as few unnecessary components as possible.

\section{FLOPs estimation for decomposition runs}
\label{app:flops}

As a rough estimate of the total FLOPs used for different decompositions, matrix multiplications are counted with the standard $2mnk$ rule. 
Attention is counted as $4\times\text{sequence length}\times d_\text{model}$ FLOPs per token per layer (query–key and attention–value products); unembedding as $2\times d_\text{model}\times \text{vocab size}$ FLOPs per token. Softmax, RMS-norm and activation functions are neglected.
Each training step on a batch of $B$ sequences of length $L$ involves:
\begin{itemize}
    \item One forward pass through the frozen target model;
    \item One forward and one backward pass through the causal importance network;
    \item One forward and one backward pass through the decomposed model for each of the reconstruction losses (stochastic, adversarial, unmasked),
    for both the target and non-target batches. Each decomposed linear layer has a cost of $2 C (d_\text{in} + d_\text{out})$ per token, with $C$ the number of components.
    The adversarial optimization steps are counted in the same way.
\end{itemize}
The cost of a backward pass is approximated as twice the cost of the corresponding forward pass.

\section{Does parameter-decomposition produce incomplete circuits?}
\label{app:completeness}

In all current parameter-decomposition methods, the loss functions require the decomposition to accurately reconstruct the model's output using a minimal number of components.
This could cause problems in cases where the model possesses multiple mechanisms that perform a similar function, such that only a subset of these are enough to replicate the model's output \citep{wang_interpretability_2022}.
Concretely, in transformers, this could happen when mechanisms write to the same saturated softmax function (either in attention blocks, or in the final softmax).
In such cases, PD's minimality loss would likely inactivate as many of these redundant mechanisms as possible, leading to an incomplete explanation.
This could compromise applications like unlearning, where we wish to erase all instances of a fact.

This issue has not yet been addressed even in full-data PD, but it could be particularly critical in targeted PD, where inactive mechanisms are moved to the $\Delta$ component and essentially lost.

To explore the scope of the problem in our case studies, we use a simple test:
if a subcomponent $X$ has a hidden, redundant equivalent among the inactive or $\Delta$ components, 
then the effect of ablating $X$ should be larger in the barebone circuit (active components only)
than in the full model.

\begin{figure}
    \centering
    \includegraphics[width=\linewidth]{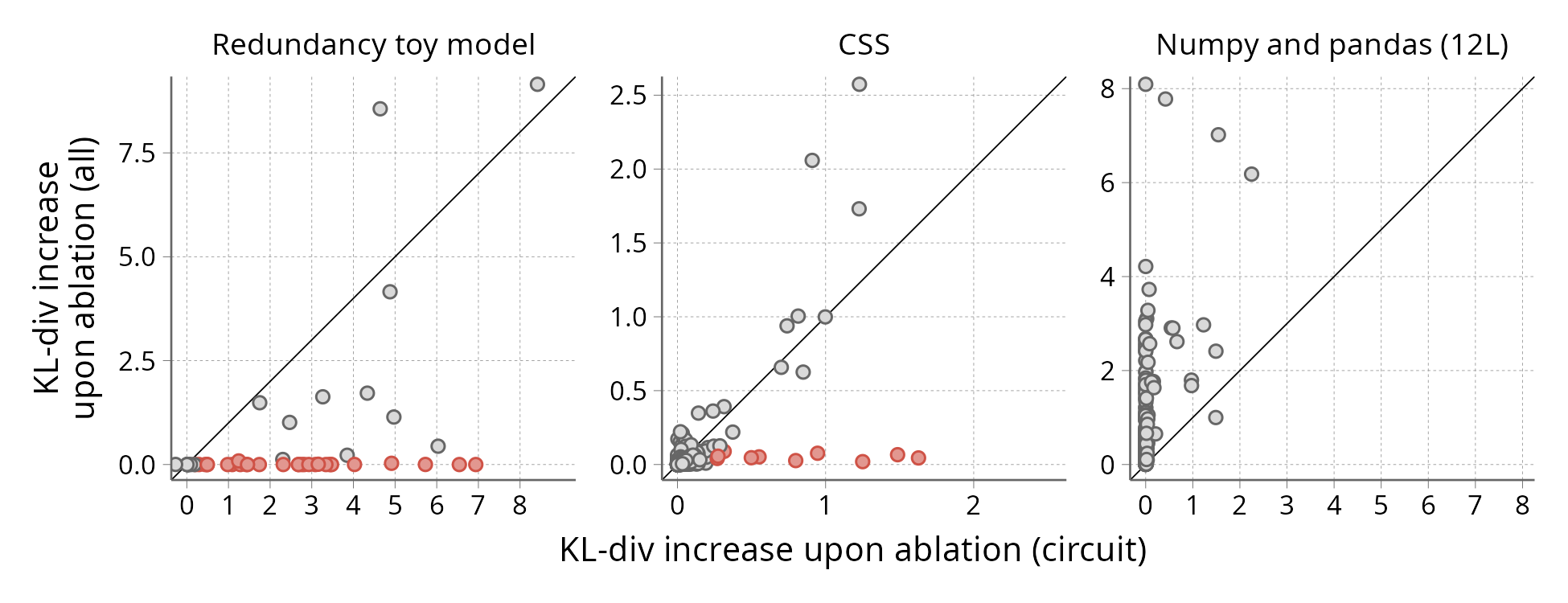}
    \caption{
        A coarse test for finding subcomponents that have a hidden redundancy among inactive/$\Delta$ components.
        The KL-divergence increase is defined as 
        $<\text{KL}(W_\text{original}x || W_\text{background without subcomponent}x) - \text{KL}(W_\text{original}x || W_\text{background}x)>$,
        where $\text{background}$ is either the full model, or the ``circuit" containing only active components from PD, 
        and the angle brackets denote the average over inputs for which this subcomponent is active.
        Subcomponents that cause a KL increase below 0.1 on the full model and above 0.2 on the decomposed circuit are highlighted in red.
    }
    \label{fig:app:completeness}
\end{figure}

However, this works reliably only under two conditions:
\begin{itemize}
    \item Ablating the component in the full model must cause near-zero degradation -- the compensation by the hidden component has to be near-total. Otherwise, the ablation derails the model's behavior and the divergences are no longer quantitatively comparable. This is visible in the ``numpy'' panel of Fig. \ref{fig:app:completeness}, where ablations are often \emph{more severe} in the original model than in the decomposed circuit.
    \item Ablating the component in the decomposed circuit must cause measurable degradation, so we can confidently assess that the same ablation has a smaller effect on the full model.
\end{itemize}

In other words, we look for components that cause a large increase in KL(original$\|$ablated) when using only the circuit (no inactive or $\Delta$ components), 
but a small increase when ablating from the full model (with inactive and $\Delta$ components on). 
This test only applies to a small fraction of the subcomponents,
but is informative as a preliminary exploration.

As a positive control, we also include a simple toy model designed to exhibit redundant mechanisms.
Given input ``$x=$" with $x\in\{1,\dots,15\}$, the model is trained to copy $x$ to position 1.
After embedding from a vocabulary of size 16 to $d_\text{model}$=64,
the model contains two single-head causal self-attention layers, no MLP or LayerNorm, followed by a non-residual linear (64 to 64) projection, unembedding and softmax.

During training (5000 steps, batch size 1024, AdamW, lr $10^{-3}$ cosine decay), each attention layer's residual
contribution is independently dropped with probability 0.4 (keeping at least
one).
This forces the model to learn the copying task twice (once per attention block), 
with the final softmax acting as an OR gate so that either block alone suffices.
The final projection is always required, as an example of a mechanism with no redundant equivalent.

Fig.\ref{fig:app:completeness} shows the results on the redundancy toy model, the CSS-only decomposition and 12-block numpy/pandas decomposition.
For the redundancy toy model, the decomposition misses many attention subcomponents -- visible from the many subcomponents that have a redundant equivalent in the $\Delta$ components (red dots).
The numpy/pandas decomposition appears to capture all relevant computation, but there is preliminary evidence that the CSS-only decomposition misses some redundant mechanisms hidden in the $\Delta$ components.

\section{Hyperparameters and data preparation}

\paragraph{4-blocks numpy/pandas decompositions}
The target data consists of short prompts of three tokens: \texttt{import numpy as}, \texttt{import pandas as}, or both.
The non-target data is sampled from the Pile with sequence length $64$.

The training objective combines three losses:
a stochastic-subset reconstruction loss (coefficient $1$),
a persistent PGD reconstruction loss (coefficient $0.5$, activated after $80\%$ of training, with a cosine inner learning-rate schedule),
and an importance-minimality penalty whose strength and p-norm vary throughout training.
The p-norm is annealed linearly from $2.0$ down to $1.0$ over the full run,
while the importance-minimality coefficient is warmed up linearly from $0$ to $4\cdot 10^{-3}$ during the first $20\%$ of training and then decreased linearly to $10^{-3}$ over the remaining $80\%$.

The causal-importance network is a global shared MLP with a single hidden layer of size $512$ and uses continuous sampling.
We allocate $C=96$ components to each of \texttt{c\_fc}, \texttt{down\_proj}, and \texttt{o\_proj}, and $C=64$ components to each of \texttt{q\_proj}, \texttt{k\_proj}, and \texttt{v\_proj}.
Training lasts for $30{,}000$ steps with batch size $256$ at a $10^{-4}$ learning rate.

\paragraph{12-blocks numpy/pandas decomposition.}
The target and non-target data are identical to the 4-block setup.
The losses are also the same, with one difference:
the persistent PGD term is activated earlier, after $20\%$ of training.
The causal-importance network and component counts match the 4L configuration.
Training lasts for $30{,}000$ steps with batch size $128$ and a learning rate of $5\cdot 10^{-4}$.

\paragraph{CSS decomposition.}
For target data, we use a custom subset of the Pile with only CSS code and without comments.
CSS documents are identified via the language metadata in \verb|andstor/the_pile_github|
(selecting only the \texttt{CSS} parquet shards), and block comments are stripped with the regex \verb|/\*.*?\*/| (non-greedy, dotall) before tokenization.
The non-target data is drawn from the Pile. 
Both use sequences of length 512.

Four losses are active during training:
a stochastic-subset reconstruction loss (coefficient $1$),
a persistent PGD reconstruction loss (coefficient $1$, activated after $20\%$ of training, with a constant inner learning-rate schedule),
an unmasked reconstruction loss (coefficient $0.2$),
and an importance-minimality penalty.
The p-norm of the latter is held at $2.0$ for the first $10\%$ of training and then annealed linearly to $0.4$ over the remaining $90\%$;
its coefficient is warmed up linearly from $0$ to $3\cdot 10^{-3}$ over the first $10\%$ of training and then decreased linearly to $3\cdot 10^{-4}$ over the remaining $90\%$.

The causal-importance network is a global shared transformer with $d_{\text{model}}=1024$, $2$ blocks, an MLP hidden dim of $2048$, $4$ attention heads, a maximum length of $512$, and a RoPE base of $10{,}000$;
sampling is binomial rather than continuous.
We allocate $C=400$ components to \texttt{c\_fc}, $512$ to \texttt{down\_proj}, $126$ to \texttt{q\_proj}, $96$ to \texttt{k\_proj}, and $256$ to each of \texttt{v\_proj} and \texttt{o\_proj}.
Training runs for $50{,}000$ steps with batch size $64$ and a learning rate of $10^{-4}$.

\iftrue

    \begin{figure}[p]
        \centering
        \includegraphics[width=.75\linewidth]{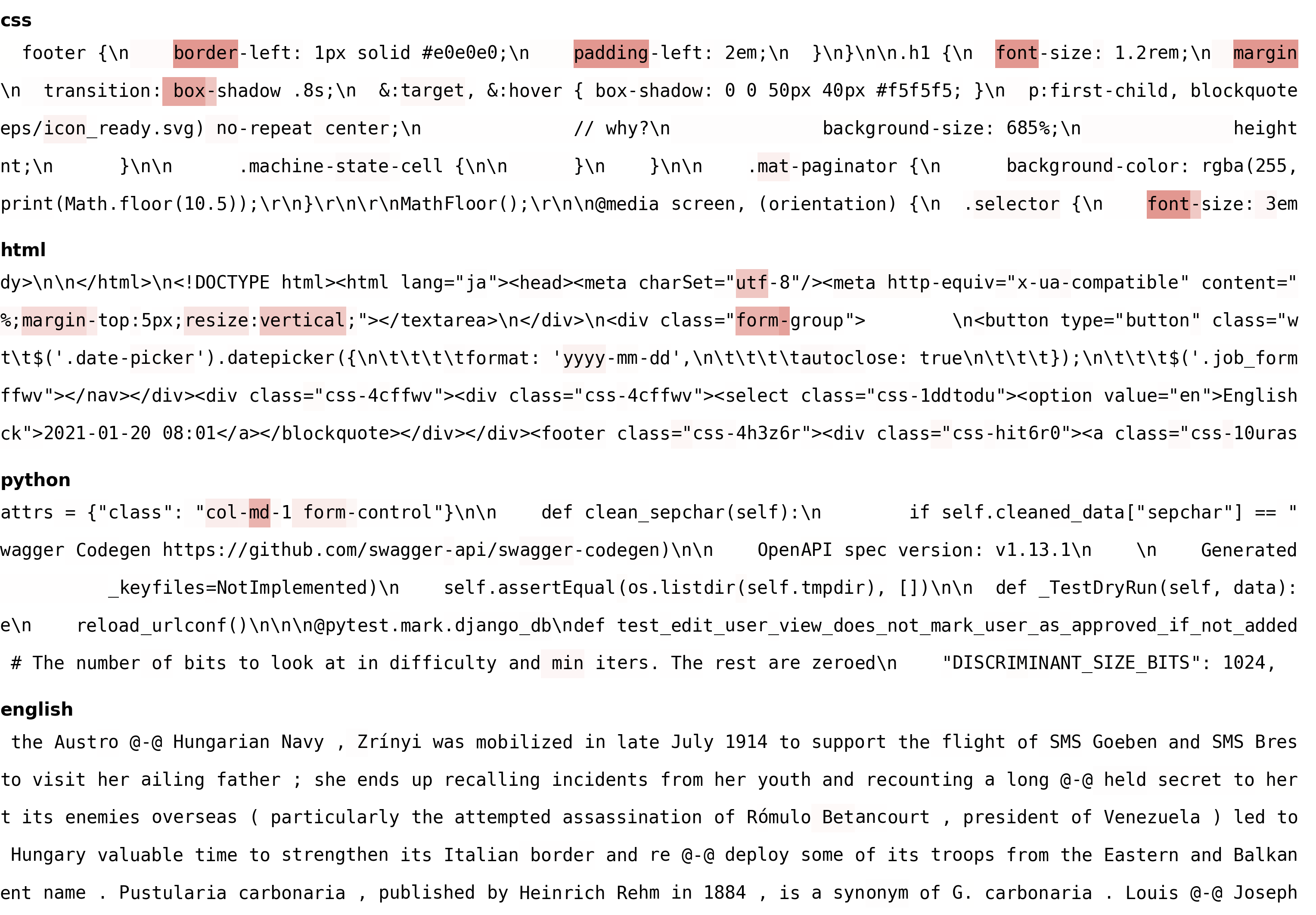}
        \caption{
            Effect of ablating \textbf{subcomponent 407 from layer 0's MLP down projection} on output reconstruction, 
            for sample input sequences of CSS code, HTML code, Python code and plain English.
            This subcomponent appears to be involved in predicting the character "-" after a keyword in CSS code.
            Incidentally, it activates in HTML and Python examples, but this happens when the examples contain snippets of CSS code.
            The shaded area represents the KL-divergence w.r.t. the original model when this component is ablated, clamped to $[0; 1]$.
            The sequences were selected by running the ablated model on 100 random inputs from the Stack \citep{kocetkov_stack_2022} and picking the top-5 sequences where the component activated the most often.
        }
        \label{fig:app:css_eoa_407}
    \end{figure}
    
    \begin{figure}[p]
        \centering
        \includegraphics[width=.75\linewidth]{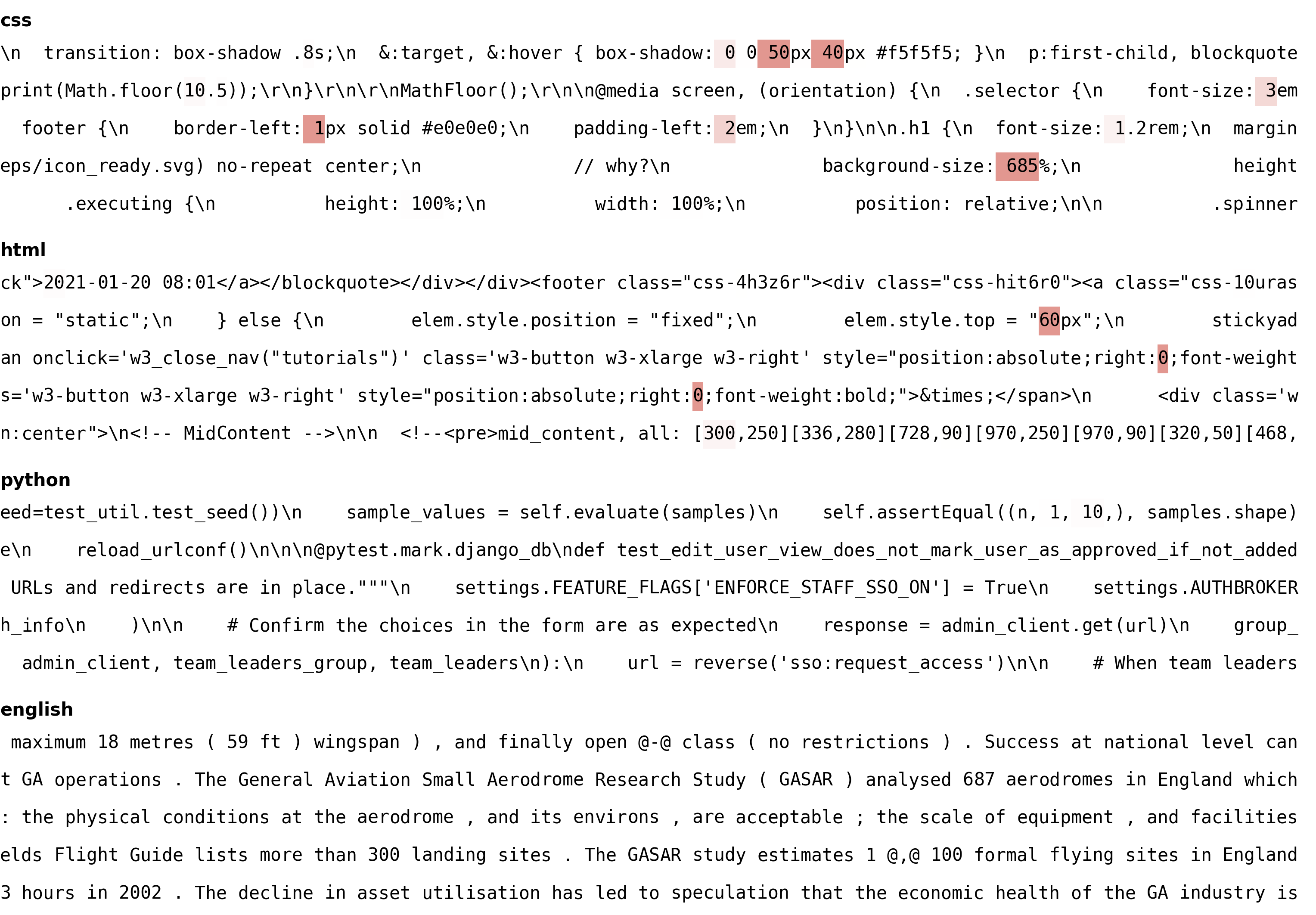}
        \caption{
            Effect of ablating \textbf{subcomponent 363 from layer 3's MLP down projection} on output reconstruction, 
            for sample input sequences of CSS code, HTML code, Python code and plain English.
            This subcomponent appears to be involved in predicting units ("px", "em", "\%") after numbers in CSS code
            The shaded area represents the KL-divergence w.r.t. the original model when this component is ablated, clamped to $[0; 1]$.
            The sequences were selected by running the ablated model on 100 random inputs from the Stack \citep{kocetkov_stack_2022} and picking the top-5 sequences where the component activated the most often.        
        }
        \label{fig:app:css_eoa_363}
    \end{figure}
    
    \begin{figure}[p]
        \centering
        \includegraphics[width=.75\linewidth]{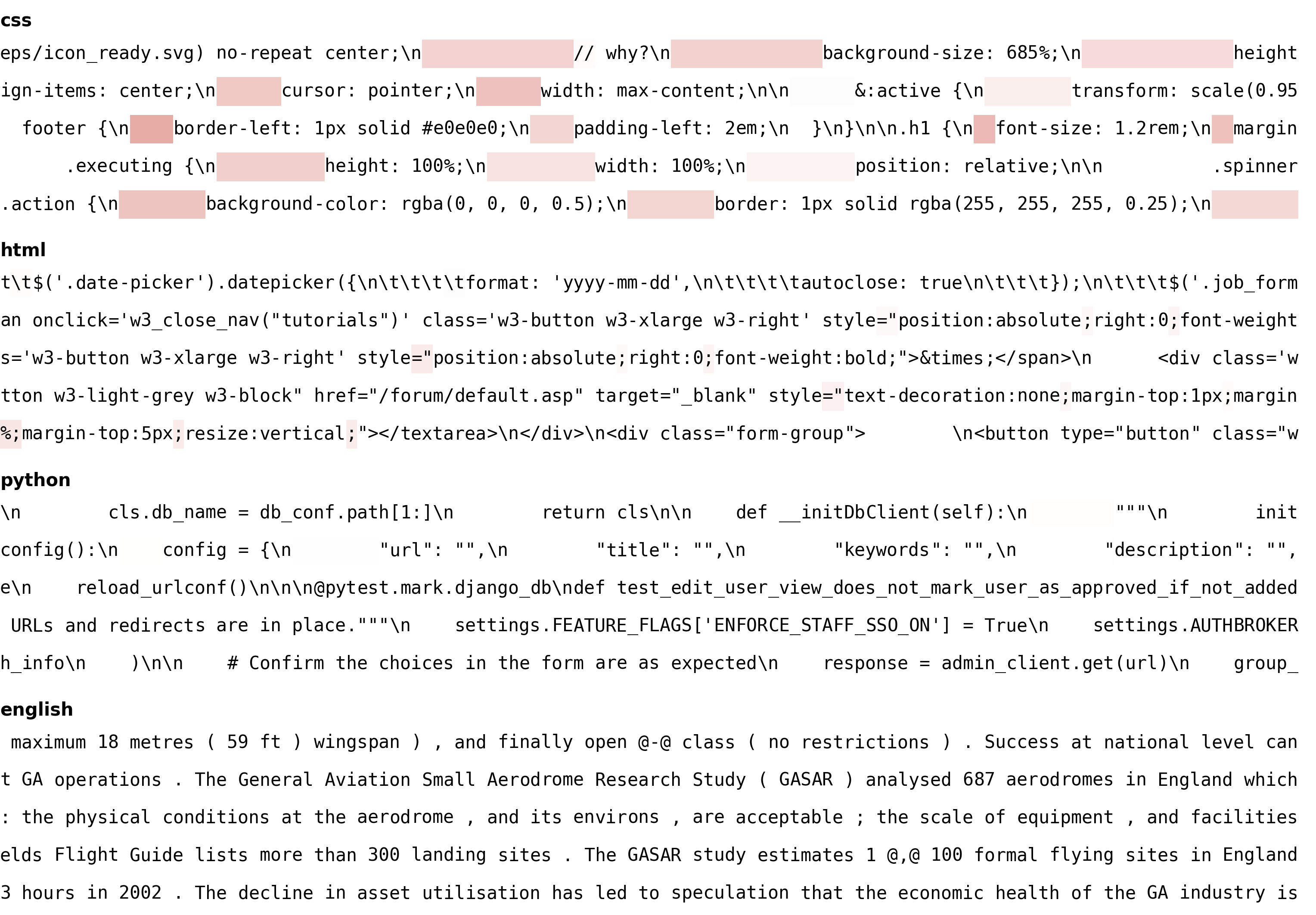}
        \caption{
            Effect of ablating \textbf{subcomponent 98 from layer 3's MLP down projection} on output reconstruction, 
            for sample input sequences of CSS code, HTML code, Python code and plain English.
            This subcomponent is active on indentation blocks in CSS code. Notably, it is not active on indentation blocks in other contexts, such as Python code.
            The shaded area represents the KL-divergence w.r.t. the original model when this component is ablated, clamped to $[0; 1]$.
            The sequences were selected by running the ablated model on 100 random inputs from the Stack \citep{kocetkov_stack_2022} and picking the top-5 sequences where the component activated the most often.     
        }
        \label{fig:app:css_eoa_98}
    \end{figure}

\fi

\end{document}